\documentclass[aps,twocolumn,superscriptaddress,amsmath,amssymb,floatfix]{revtex4-1}

\usepackage{graphicx}
\usepackage{dcolumn}
\usepackage{bm}
\usepackage{textcomp} 
\usepackage[caption=false]{subfig}
\usepackage{tabularx}
\usepackage{multirow}
\usepackage{array}

\newcommand{\fref}[1] {Figure~\ref{#1}} 
\newcommand{\tref}[1] {Table~\ref{#1}} 
\newcommand{\eref}[1] {Equation~\ref{#1}} 
\newcommand{\code}[1] {\texttt{#1}} 
\newcolumntype{L}[1]{>{\raggedright\arraybackslash}p{#1cm}}
\begin{document}

\title{A watershed-based algorithm to segment and classify \protect\\ cells in fluorescence microscopy images}

\author{Lena R. Bartell}
\affiliation{School of Applied and Engineering Physics, Cornell University, Ithaca, NY}

\author{Lawrence J. Bonassar}
\affiliation{Meining School of Biomedical Engineering, Cornell University, Ithaca, NY}
\affiliation{Sibley School of Mechanical and Aerospace Engineering, Cornell University, Ithaca, NY}

\author{Itai Cohen}
\affiliation{Department of Physics, Cornell University, Ithaca, NY}

\date{\today}

\begin{abstract}
Imaging assays of cellular function, especially those using fluorescent stains, are ubiquitous in the biological and medical sciences. Despite advances in computer vision, such images are often analyzed using only manual or rudimentary automated processes. Watershed-based segmentation is an effective technique for identifying objects in images; it outperforms commonly used image analysis methods, but requires familiarity with computer-vision techniques to be applied successfully. In this report, we present and implement a watershed-based image analysis and classification algorithm in a GUI, enabling a broad set of users to easily understand the algorithm and adjust the parameters to their specific needs. As an example, we implement this algorithm to find and classify cells in a complex imaging assay for mitochondrial function. In a second example, we demonstrate a workflow using manual comparisons and receiver operator characteristics to optimize the algorithm parameters for finding live and dead cells in a standard viability assay. Overall, this watershed-based algorithm is more advanced than traditional thresholding and can produce optimized, automated results. By incorporating associated pre-processing steps in the GUI, the algorithm is also easily adjusted, rendering it user-friendly.
\end{abstract}

\maketitle

\section{\label{sec:intro}Introduction}
In the biological sciences, fluorescence microscopy images are ubiquitous, especially for visualizing and assessing groups of cells, \textit{in situ} or otherwise. For example, a common imaging assay used throughout biology and medicine is the viability staining assay, which highlights individual cells as either alive or dead based on the color of their fluorescence~\cite{poole_detection_1996, weston_new_1990, rotman_membrane_1966, decherchi_dual_1997, lewis_cell_2003, hembree_viability_2007, ewers_extent_2001, krueger_extent_2003}. However, when such images are quantified, the fraction of viable cells is commonly computed either by manual counting or using a simple threshold and region-counting procedure~\cite{hembree_viability_2007, ewers_extent_2001, krueger_extent_2003}. Such results are useful, but the growing use of computer vision and advanced processing algorithms has opened the door to advanced, automated pipelines~\cite{wollman_high_2007}. Such analysis pipelines usually include a segmentation process to identify cells or other regions of interest, followed by an analysis and possible classification process to extract data of interest.

Segmentation is the process of identifying objects in an image (e.g. cells), thus separating them from the background and each other. Once obtained, a segmentation result enables the researcher to not only count the number of cells but also analyze individual cells to extract relevant information about them, such as their shape, size, or behavior. Automatic segmentation is more appealing than manual counting because it enables high-throughput data analysis and thus higher statistical power with increased efficiency and less human bias. A segmentation result further allows each object or cell to be analyzed based on the local image data in that region, such as staining intensity, which may reflect useful information about the cell. Thus, by identifying cell regions, segmentation enables a range of automated data analysis techniques, such as classification into different groups (e.g. alive vs dead). This type of data analysis pipeline similarly enables both spatial and temporal analysis - tracking individual cells' locations and their states over time. Without automated analysis, these methods are not applicable and their benefits are unrealized. 

The watershed segmentation algorithm has been successfully implemented to automatically find and analyze cells in fluorescence micrographs~\cite{yang_nuclei_2006}. The watershed algorithm considers the 2D grayscale image as a topographical map with “mountains” of high pixel values and “valleys” of low pixel values~\cite{moga_parallel_1998, vincent_watersheds_1991, meyer_morphological_1990}. In this analogy, the algorithm adds “water” to the image landscape to determine the independent catchment basins and the ridge lines separating those basins. Each catchment basin is then a segmented object. This algorithm is advantageous for separating objects that are close or “touching” in the image and is less susceptible to overall intensity variations~\cite{meyer_morphological_1990}. However, such watershed segmentation has resisted widespread adoption, in part because it is prone to oversegmentation, producing too many regions that are too small. Common methods to avoid oversegmentation include more advanced region-growing procedures and marker-guided segmentation~\cite{moga_parallel_1998, meyer_morphological_1990, bleau_watershed-based_2000, bieniecki_oversegmentation_2004}. However, these techniques are not readly accessible to the broad research community without advanced experience in computer vision and a scripting language. 

Thus, to make a watershed-based image analysis pipeline more accessible to biomedical researchers, we present and implement an image analysis and classification algorithm. We deploy this algorithm together with associated pre- and post-processing steps in a graphical user interface (GUI), enabling the user to easily understand and adjust the various steps to their specific needs. As an example, we implement this algorithm to find and classify cells in a three-color fluoresce microscopy assay of mitochondrial function where the cells have complex, non-contiguous staining. As a second example, we apply the algorithm to a two-color, live-dead staining assay and outline a workflow to optimize and validate the parameters using manual identification and receiver operator characteristics. This algorithm, which relies on watershed-based segmentation, is more advanced than traditional thresholding and can thus provide more reliable results. By incorporating associated pre-processing steps in the GUI, the algorithm is also easily adjusted. Similarly, by incorporating post-processing analyses, we enable the user to extract additional data of interest.


\section{\label{sec:algorithm}Algorithm}
The algorithm presented here includes image filtering and segmentation followed by object classification. For segmentation, this algorithm first filters the image then prepares for and applies background-enforced watershed segmentation, with optional realistic limits on the results. After segmentation, the resulting objects can be classified into two groups by thresholding based on their pixel intensities.

\subsection{\label{sec:methods-segmentation}Segmentation}
Before segmentation, the raw image is filtered to reduce common flaws and artifacts and modified to prepare for segmentation. First, if the raw image is a 3-color channel (i.e. RBG) image, it is converted to grayscale intensity image using the MATLAB function \code{rgb2gray} (The MathWorks Inc., Natick, MA). All subsequent segmentation operations are performed on a grayscale image. Second, adaptive histogram equalization (\code{adapthisteq}) is used to reduce the effects of uneven illumination. The degree of illumination correction and contrast enhancement is controlled by the \textit{equalization clip limit} parameter, which ranges from 0 to 1, with a default value of 0.01. Third, a median filter (\code{medfilt2}) is used to compute the image background, which is then subtracted from the image. This background-subtraction step reduces the effects of out-of-plane signal and uneven illumination, and increases the contrast between objects and their adjacent background. The size of the median filter is governed by the \textit{background size} parameter.  This size, which specifies the diameter of the median filter, should be an odd integer that is larger than any object you are trying to segment. Fourth, a second median filter (\code{medfilt2}) and a Gaussian filter (\code{imfilt}) are applied to reduce high-frequency components in the image.  The diameter of this second median filter is governed by the \textit{median size} parameter while the Gaussian radius is governed by the \textit{Gaussian radius} parameter. The \textit{median size} should also be an odd integer approximately equal to or smaller than the objects to be segmented. The \textit{Gaussian radius} can by any positive real number but should also be smaller than the objects to be segmented. These median and Gaussian filters serve both to reduce pixel-level noise and to smooth and blur away small-scale image variations within an single object . Combined, these four steps produce a filtered image that is used for watershed segmentation. In the GUI the latter three steps are optional and can be selectively included or excluded from the filtering process if desired. If a given step is not necessary to achieve a good segmentation result, exclusion may be beneficial to save computation time, for example.

To segment objects, background-enforced watershed segmentation is applied. This is a variant of watershed segmentation where some pixels in the image are forced to be part of the background (i.e. peaks or valleys).  First, the filtered image is thresholded using a two-level Otsu filter, breaking the image into three groups (\code{multithresh})~\cite{otsu_threshold_1979}. Pixels in the darkest group are said to belong to the watershed background.  Second, the filtered image is inverted so objects are dark on a light background, i.e. objects are catchment basins.  Third, all background pixels are then set to the same low elevation, so they will collect into one catchment basin that is later discarded. Fourth, the watershed transform is applied (\code{watershed}), returning a segmentation result that defines unique catchment basins. Then, all segmented objects that contain one or more background pixels are forced to be part of the background and thus discarded. Finally, realistic limitations on object size and brightness are enforced. In particular, the \textit{minimum area} and \textit{maximum area} parameters, specified in square pixels, impose limitations on the area of any segmented object in the image. Similarly, the \textit{minimum signal} parameter specifies the minimum average intensity within an object, such that very dim segmented objects are discarded. The minimum signal is specified as a number ranging from 0 to 1, representing the full range of intensities in the raw grayscale image.

The result of the filtering and segmentation steps is a segmentation result, where each pixel in the image is identified as belonging to the background or the first object or the second object, etc. This information is stored as a label matrix the same size as the original image, where the value at each pixel specifies the object that pixel belongs to (0 is background, 1 is the first object, etc.). This segmentation can be linked back to the raw intensity image and used for post processing and to compute outcomes of interest for each object, such as the object’s position, area, and average intensity in each color channel. In particular, the raw image and segmentation result are used to classify the found objects.

\subsection{\label{sec:methods-classification}Classification}
After segmentation is complete, the result is related back to the raw image in order to classify objects based on the pixel intensities in each color channel. Each object, as defined by the segmentation result, contains many pixels. In the raw image, these same pixels are associated with pixel intensities, where each pixel has three intensity values associated with it: red, green, and blue (in that order, by default). For example, if segmented object number 15 contains 121 pixels (perhaps it is an 11 pixel $\times$ 11 pixel square), then there are 3 colors $\times$ 121 pixels or 363 intensity values associated with that object. These intensity values are utilized to classify each segmented object.

To classify objects, a classification function $f(R,G,B)$ is applied to each object, where $R$, $G$, and $B$ are the list of red, green, and blue pixel values in a given object, respectively. As such, this MATLAB-based function $f$ should take three lists of equal length and return one scalar value output. For example, the function \code{mean(R)-mean(G)} subtracts the mean value of green pixels from that of red pixels, while blue values are ignored; if the result is greater than zero then there is more red signal in the object than green signal, and vice versa. In the GUI, this function $f(R,G,B)$ can be varied to meet the users' needs. When applied, the function produces one scalar value for each segmented object. The GUI displays a histogram of theses scalar outputs. After calculating the classification function, the resulting scalar values are thresholded to divide the associated objects between two classes, or states: objects with scalar values above the threshold are considered to be in state 1 and those below in state 2. The threshold value can be chosen either manually or automatically via Otsu's method~\cite{otsu_threshold_1979}. The scientific interpretation of these two states will depend on the imaging setup (e.g. fluorescent stains, imaging filters), the design of the classification function, and the threshold value.  Note that, although only three-color raw images are discussed above, this process is equally valid for grayscale images. Grayscale images are represented in three-color (red, green, and blue) but with all color channels equal, resulting in a simple grayscale intensity variation. 
\subsection{\label{sec:methods-Code}Code}
The algorithm and associated MATLAB-based GUI are freely available online, along with instructions for installation and use: \href{https://github.com/itaicohengroup/watershed_cells_gui}{https://github.com/itaicohengroup/watershed\_cells\_gui}.

\section{\label{sec:example}Example implementation}
To illustrate this algorithm, we apply it to an example microscopy image and show the output at each step.

\begin{figure*}[!htb]
    \centering
    \includegraphics[width=\textwidth]{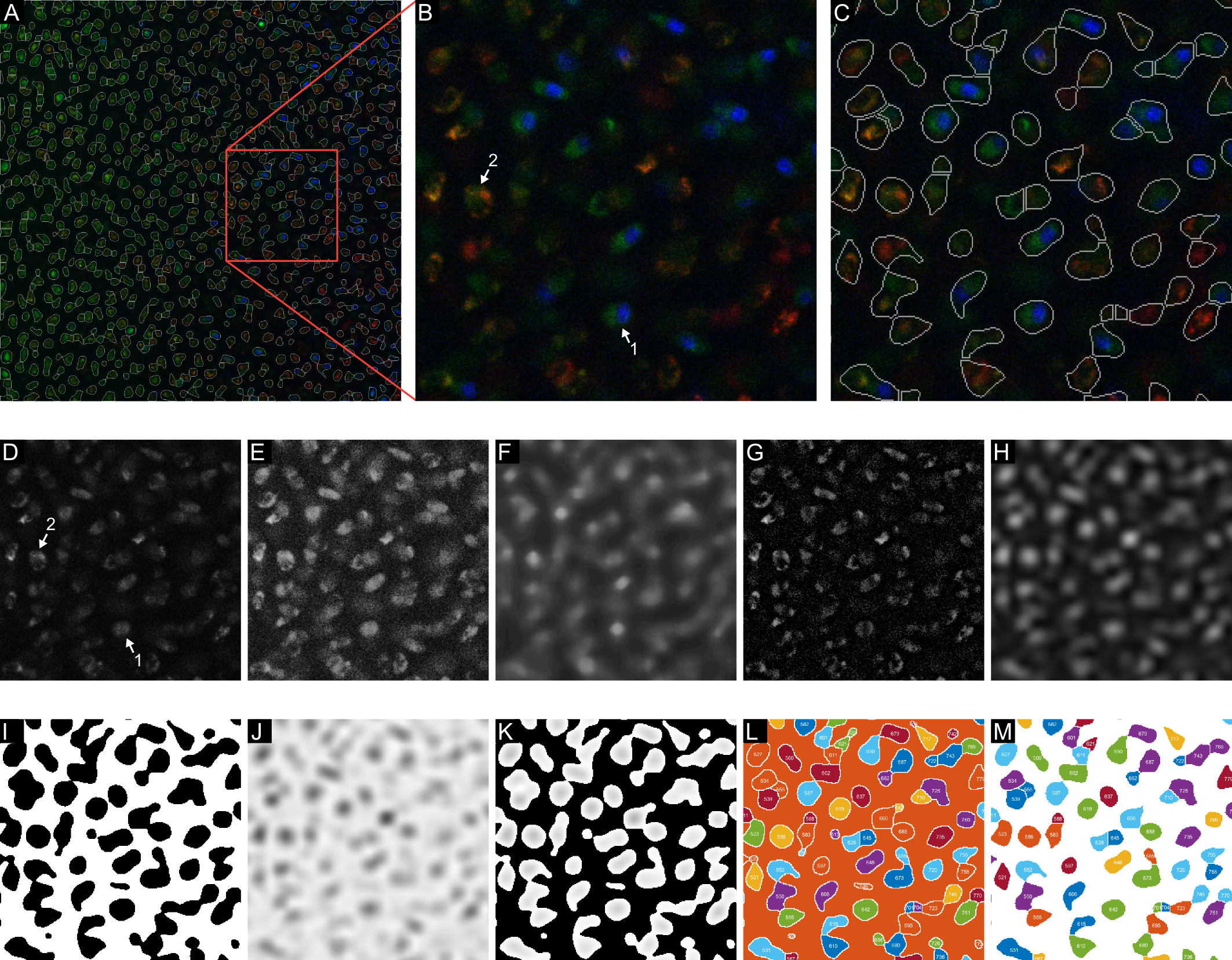}
    \caption{Example implementation of filtering and segmentation steps (as detailed in the text and \tref{tab:steps}).  Frame (A) shows the full image with segmented regions outlined in gray, where the red square highlights a subset of the full image that is explored for the remainder of this figure, including (B) the raw image subset and (C) the corresponding final segmentation result. Filtering steps1-4 and segmentation steps 1-5 are shown for this subset in (D-H) and (I-M), respectively. The subset is shown after (D) grayscale conversion and (E) histogram equalization. The background image (F), is subtracted from (E) to produce (G), which is subsequently smoothed, producing (H). The binary background is shown in (I), with the background in white, and (J) shows the inverted filtered image. The background is then enforced, yielding (K). Finally, watershed segmentation on (K) produces the numbered segmented regions indicated in (L), and unrealistic regions are discarded (as is the background region), producing the  regions shown in (M). The segmented regions in (M) are the same regions outlined in (A) and (C).}
    \label{fig:implementation}
\end{figure*}

\subsection{\label{sec:example-prep}Sample preparation}
The example image (\fref{fig:implementation}) shows chondrocytes in a neonatal bovine articular cartilage tissue explant. The explant was dissected sterilely from the medial condyle of a neonatal bovine stifle joint (sex unknown). The 6 mm-diameter explant was rinsed in phosphate buffered saline, incubated overnight, and bisected to create a hemicylinder. One hemicylinder was stained for 50 minutes in 200 nM MitoTracker Green to stain all of the mitochondria green, for 20~minutes in 10~nM Tetramethylrhodamine methyl ester (TMRM) to stain only polarized mitochondria red, and 20~minutes in 100~nM Sytox Blue to stain all permeable nuclei blue (all stains: ThermoFisher Scientific, Waltham, MA). Thus, the red fluorescence highlights only polarized mitochondria, the green fluorescence highlights all mitochondria, and the blue fluorescence highlights dead cells. Images of stained chondrocytes in the whole tissue explant were collected using a confocal microscope (LSM 880 inverted, Carl Zeiss Microscopy, Jena, Germany). An image was taken of the flat cut surface using a 20$\times$ objective. 

\begingroup
\squeezetable
\setlength{\tabcolsep}{6pt}
\begin{table*}[!htt]
    \centering
    \begin{ruledtabular}
    \begin{tabular}{L{1.3} L{1.3} L{0.8} L{3.8} L{3.8} L{2.8}}
        \multicolumn{2}{l}{\textbf{Stage}} & \textbf{Step} & \textbf{Description} & \textbf{Parameter} & \textbf{Value} \\ 
        \hline
        Segmentation   & Filter       &   1 & Convert to grayscale    &                           &               \\
                       &              &   2 & Histogram equalization  & Equalization clip limit   & 0.01          \\
                       &              &   3 & Background subtraction  & Background size           & 19 px         \\
                       &              &   4 & Smoothing               & Median size               & 7 px          \\
                       &              &     &                         & Gaussian radius           & 7 px          \\
        \cline{2-6}
                       & Watershed    &   1 & Determine background    &                           &               \\
                       &              &   2 & Invert                  &                           &               \\
                       &              &   3 & Enforce background      &                           &               \\
                       &              &   4 & Watershed               &                           &               \\
                       &              &   5 & Impose realistic limits & Minimum area              & 35 px$^2$     \\
                       &              &     &                         & Maximum area              & 2000 px$^2$   \\
                       &              &     &                         & Minimum signal            & 0.2           \\
        \hline
        Classification &              &   1 & Compute scalar function for each segmented object   & Function $f(R,G,B)$       & \code{mean(R)} \\
                       &              &   2 & Threshold scalar output to determine state          & Threshold                 & $0.35 = 9/255$ \\
    \end{tabular}
    \end{ruledtabular}
    \caption{Steps in the segmentation and classification algorithm, their associated parameter names, and the parameter values applied to analyze the example image shown in \fref{fig:implementation} and \fref{fig:classification}.}
    \label{tab:steps}
\end{table*}
\endgroup

\subsection{\label{sec:example-segmentation}Segmentation}
The example image of the mitochondrial function assay was analyzed using the steps described above, implementing the parameters as detailed in \tref{tab:steps}. Example step-wise outputs of filtering and segmentation are illustrated in \fref{fig:implementation}. \fref{fig:implementation}A-C shows the raw image and segmentation result. \fref{fig:implementation}D-H illustrates the filtering steps 1-4 applied to this image, and \fref{fig:implementation}I-M shows output from segmentation steps 1-5. Note that, in the raw and grayscale image (\fref{fig:implementation}A-B,D), cells do not have consistent staining morphology. Due to the nature of the stains used, some cells have relatively constant pixel values throughout (arrow 1), while others have speckled staining pattern in an annular shape (arrow 2). This pattern results because a cell generally contains multiple mitochondria which all stain red with TMRM but are excluded from the nucleus. Similarly, the relative intensity of staining varies between cells. Each of these factors complicates and reduces the ability of simple thresholding to accurately identify and segment regions. 

\begin{figure}[!ht]
    \centering
    \includegraphics[width=\columnwidth]{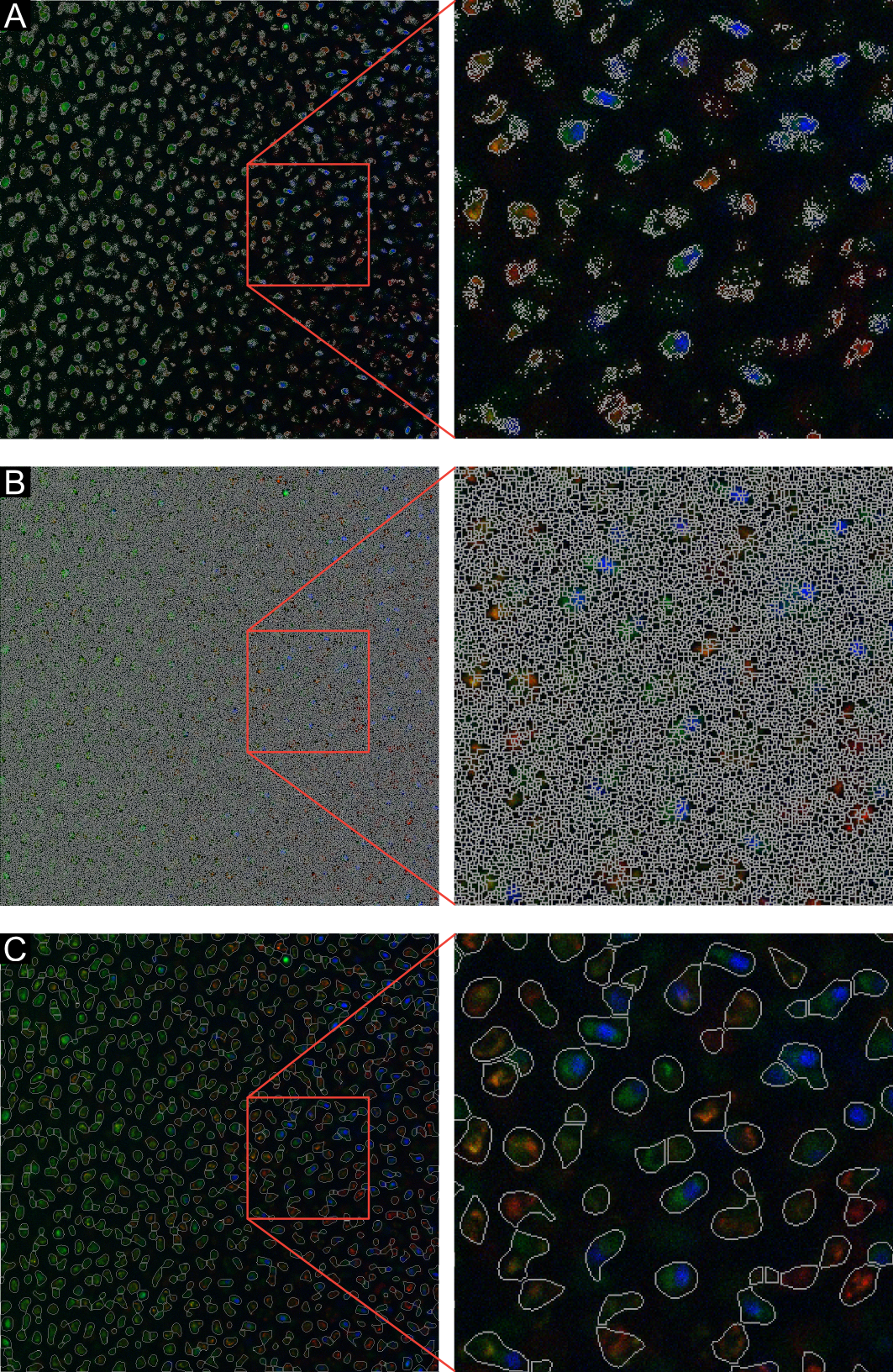}
    \caption{Comparison of segmentation results obtained by applying different methods to the grayscale image. In each frame, segmented objects are outlined in gray. Segmentation is compared for (A) Otsu thresholding, (B) watershed segmentation after inversion but without any filtering or pre-processing, and (C) the custom watershed algorithm described in this manuscript.}
    \label{fig:comparison}
\end{figure}

\fref{fig:comparison} shows the segmentation results obtained by this custom watershed-based algorithm compared to segmentation by Otsu thresholding and by watershed segmentation without any filtering or pre-processing beyond inversion. Both threshold and simple watershed methods display over-segmentation, with much more in the latter. Thresholding (\fref{fig:comparison}A) is sensitive to intensity variations across the image, resulting in smaller, poorly segmented regions in the bottom right corner of the full-scale image. Watershed segmentation with appropriate filtering results in gross oversegmentation (\fref{fig:comparison}B). The custom filter and watershed-based method described in this manuscript (\fref{fig:comparison}C) is imperfect but a clear improvement. The custom watershed method consistently segments cells across the image and separates most adjoining cells, though there are some errors from erroneously separating or combing neighboring cells.

\begin{figure}[!ht]
    \centering
    \includegraphics[width=\columnwidth]{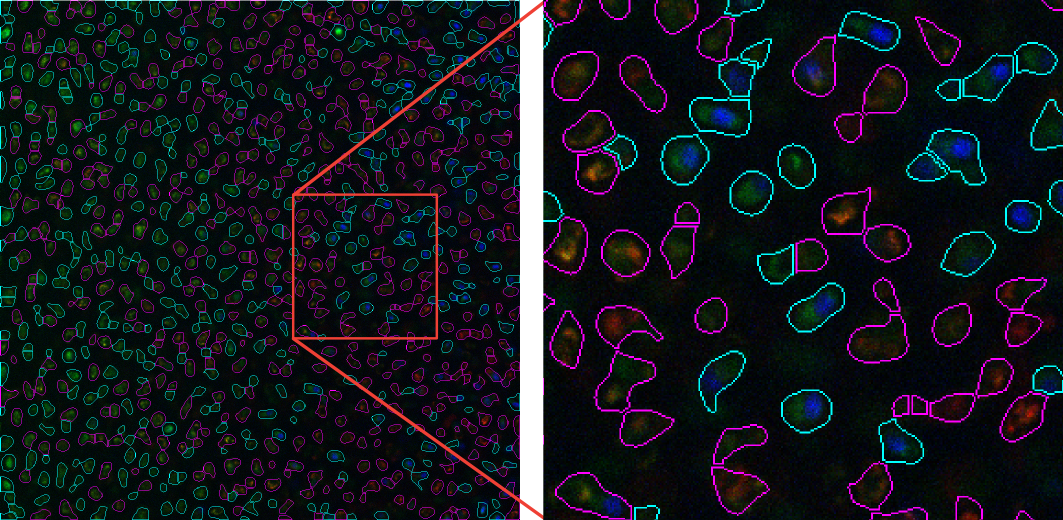}
    \caption{Classification output for the example image, shown in full (left) and for a subset (right). Cells in state~1 (polarized mitochondria) are outlined in magenta and cells in state~2 (depolarized mitochondria) are outlined in cyan.}
    \label{fig:classification}
\end{figure}

\begingroup
\squeezetable
\setlength{\tabcolsep}{3pt}
\begin{table}[!ht]
    \centering
    \begin{ruledtabular}
    \begin{tabular}{c c c c c c c}
                       & \multicolumn{3}{c}{\textbf{Cell A}}  & \multicolumn{3}{c}{\textbf{Cell B}}   \\
        Pixel \#       & Red       & Green     & Blue          & Red       & Green     & Blue          \\
        \cline{1-1} \cline{2-4} \cline{5-7}
        1              &  0        & 12        & 28            &  0        & 21        &  5            \\
        2              &  6        &  6        & 19            &  9        & 30        &  2            \\
        3              & 20        & 18        &  5            &  0        & 15        & 15            \\
        4              &  0        &  9        & 15            &  5        & 13        &  3            \\
        5              & 13        & 12        &  2            &  8        & 13        & 11            \\
        6              &  0        & 12        & 28            &  0        & 21        &  5            \\
        $\vdots$       & $\vdots$  & $\vdots$  & $\vdots$      & $\vdots$  & $\vdots$  & $\vdots$      \\
        Mean           & 17        & 10        &  9            &  3        & 16        & 44            \\
        \cline{1-1} \cline{2-4} \cline{5-7}
        Classification & \multicolumn{3}{c}{State 1}          & \multicolumn{3}{c}{State 2}           \\
    \end{tabular}
    \end{ruledtabular}
    \caption{Two regions, or cells, chosen from the example image in \fref{fig:classification}, and a subset of their associated pixel values in the red, green, and blue channels. The mean pixel values are also shown, indicating that, based on the parameters in \tref{tab:steps}, cell~A would fall into state~1, and cell~B into state~2.}
    \label{tab:pixel_values}
\end{table}
\endgroup

\subsection{\label{sec:example-classification}Classification}
After segmentation, the example image of the mitochondrial function assay was analyzed to classify cells based on mitochondria polarity, using the parameters detailed in \tref{tab:steps}. In the staining assay utilized here, only cells with polarized mitochondria stain red. As such, only the red pixels from each cell were analyzed and the scalar classification function $f$ was chosen to be \code{mean(R)}. These scalar values were thresholded at the value shown in \tref{tab:steps} to classify each cell as having either polarized mitochondria (state~1) or depolarized mitochondria (state~2). The threshold value was chosen by visual inspection. As an example, \tref{tab:pixel_values} lists the red, green, and blue pixel values for two example regions and their mean values. Because the image was collected with 8~bit depth, the pixel values range from 0 to 255. In this case, cell~A has a scalar value above the threshold of 9 ($f(R,G,B)=17$) and thus is in state~1 with polarized mitochondria, while the opposite is true for cell~B. The complete results of this classification are illustrated in \fref{fig:classification}, with regions in state~1 outline in magenta and those in state~2 outlined in cyan. Based on these results, there were a total of 844 regions, with 462 in state~1 and 382 in state~2.


\section{\label{sec:optimization}Optimization workflow}
As a second example, we apply our custom watershed algorithm to an image of a standard viability assay. By comparing to manually-determined ground truth, we illustrate a protocol for optimizing the segmentation and classification parameters for a particular assay. In this example, we optimize the segmentation parameters for cell-counting procedures and optimize the classification threshold for separating live and dead cell populations.

\begin{figure}[!ht]
    \centering
    \includegraphics[width=\columnwidth]{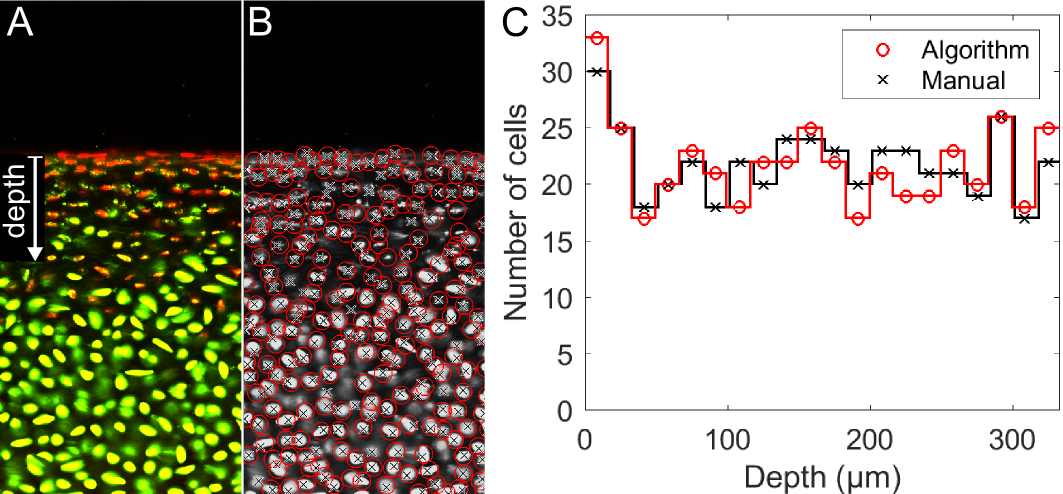}
    \caption{Viability staining assay in articular cartilage and associated optimal segmentation results. (A) Subset of raw image showing live cells in green and dead cells in red, (B) grayscale subset from (A), shown with manually-located cells’ centroids marked by black crosses and automatically found cells’ centroids indicated by red circles. (C) Binned centroids as a function of depth (as indicated in (A)), for both manually and automatically determined cells.}
    \label{fig:optimal_segmentation}
\end{figure}

\subsection{\label{sec:optimization-prep}Sample preparation}
The second example image (\fref{fig:optimal_segmentation}A) also shows chondrocytes in a neonatal bovine articular cartilage tissue explant. The explant was dissected from the femoral condyles of a neonatal bovine (sex unknown). The 6 mm-diameter explant was rinsed in phosphate buffered saline, exposed to mechanical perturbation, and then bisected to create a hemicylinder. This hemicylinder was then stained for 20~minutes in 4~µM calcein AM to stain the cytoplasm of live cells green and 2~µM ethidium homodimer to stain the nuclei of dead cells red (all stains: ThermoFisher Scientific, Waltham, MA). Thus, the red fluorescence highlights dead cells while green fluorescence highlights all live cells, revealing the amount and spatial distribution of cell viability. Images of stained chondrocytes in the whole tissue explant were collected using a confocal microscope (LSM 710 inverted, Carl Zeiss Microscopy, Jena, Germany). An image was taken of the flat cut surface of the tissue, showing the depth-profile of the tissue.

\subsection{\label{sec:optimization-segmentation}Segmentation parameter optimization}
To optimize the segmentation parameters for this viability assay, segmentation results were compared to manual counting. Custom MATLAB code was used to display the example image and collect user-selected locations of all cells in the image. Automatically computed segmentation results, particularly the label matrix, were used to compute the centroids of each segmented region or cell. In addition to visual inspection, the automatic and manual counting results were compared by binning the number of cells as a function of tissue depth and computing a sum of squared-differences metric, $w$, given by \eref{eq:w}. The algorithm parameters were varied, visually inspecting the result and computing the metric, $w$, for each set of parameters. Note that, in this case, the parameter space was explored manually, but this processes could be automated with standard or advanced parameter optimization protocols to achieve more optimal parameter convergence. The resulting optimal segmentation is shown in \fref{fig:optimal_segmentation}. Note that this procedure was designed to determine the best parameters for accurate cell counts.

\begin{equation}
    w = \sum_{bins}^{}(N_{manual} - N_{automatic})^2
    \label{eq:w}
\end{equation}

\subsection{\label{sec:optimization-classification}Classification parameter optimization}
\begin{figure}[!ht]
    \centering
    \includegraphics[width=\columnwidth]{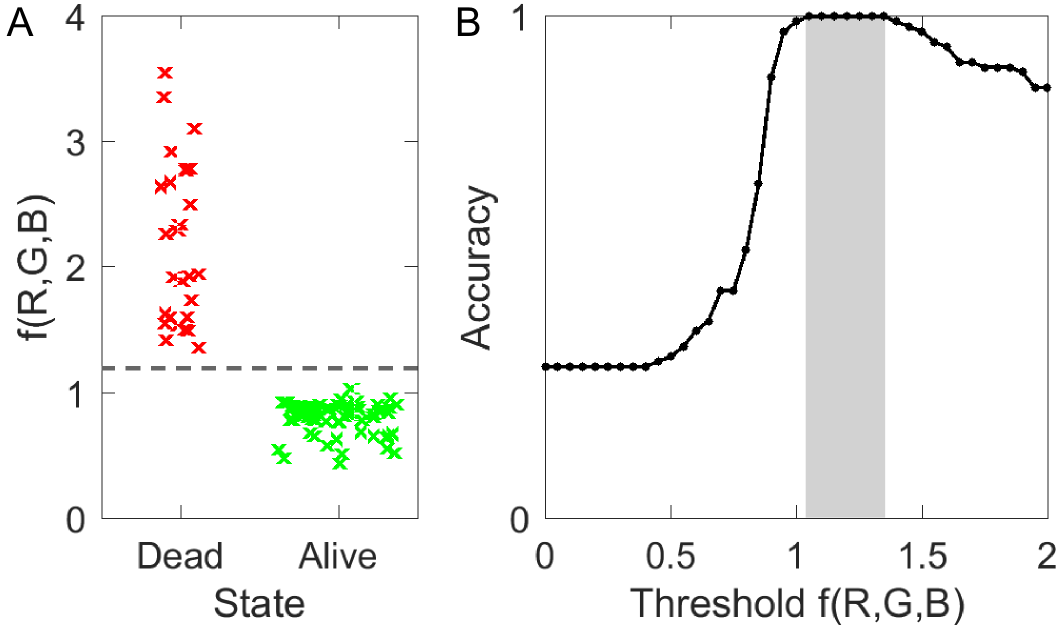}
    \caption{Optimal segmentation results for the cells segmented in the viability staining assay shown in \fref{fig:optimal_segmentation}. (A) Distribution of classification function scalar values, f(R,G,B), for a subset of 100 cells, plotted against their manually-determined state: either dead or alive. The optimal threshold of 1.2 is indicated by the dashed line.  (B) To determine the optimal threshold, i.e. the threshold with maximal accuracy, accuracy was plotted for thresholds varying from 0 to 2.}
    \label{fig:optimal_classification}
\end{figure}

The classification function was chosen based on the underlying staining mechanisms. In this assay, live cells stain green while dead cell stain red. To reflect this staining mechanism, the classification function was chosen to be \code{mean(R)/mean(G)}.  Thus, live cells will tend to have $f$ values below 1 and dead cells will tend to be above 1.
Standard receiver operating characteristic analysis methods were applied to determine the optimal threshold value of $f$~\cite{steyerberg_assessing_2010}. A randomly subset of 100 segmented cells ($\approx$20\% of all cells) was selected for manual classification. Each cell in this subset was manually determined to be either alive or dead, based on visual inspection of the image. These manual state determinations were compared to the scalar values from the classification function to determine the best threshold value. The threshold was varied from 0 to 2 and its accuracy was calculated based on true positive and true negative rates (\eref{eq:accuracy}). Based on these results, there is a range of threshold values around 1.2 with maximal accuracy, as shown in \fref{fig:optimal_classification}. As such, the optimal classification threshold was chosen to be 1.2 for this imaging assay.

\begin{equation}
    \textrm{Accuracy} = \frac{N_\textrm{True positive} + N_\textrm{True negative}}{N_\textrm{Total}}    
    \label{eq:accuracy}
\end{equation}

\section{\label{sec:discussion}Discussion}

Image analysis is an important tool used to quickly and automatically extract quantitative data from images. However, advanced image analysis techniques can be complex and non-intuitive, especially for a researcher without experience in computer vision. For this reason, despite the plethora of available techniques, researchers often analyze their images only qualitatively, or with rudimentary quantitative methods. Making advanced techniques more user-friendly can accelerate scientific understanding by enabling many researchers to extract more data without any hardware updates.

To address this need, we developed a relatively simple watershed-based algorithm for segmenting cells in 2D fluorescence microscopy images. To make the algorithm transparent yet easy to implement, we deployed this algorithm, along with pre- and post-processing steps in a MATLAB-based GUI. This algorithm and GUI consist of two primary steps: segmentation and classification. To combat the common problem of oversegmentation from the watershed algorithm, we include a variety of pre-processing filtering steps. To demonstrate its use, we applied this algorithm to two different fluorescence imaging assays: mitochondrial function and cell viability. We further optimized the parameters for segmenting and classifying cells in the viability assay, demonstrating a pipeline for other applications. 

In the literature, a variety of other techniques and software tools have been developed to similarly address the need for accessible image analysis in the biomedical sciences. For example, ImageJ includes many common image analysis protocols in a graphical interface and also allows custom plug-ins and macro-development~\cite{schneider_nih_2012}.  Other tools for image analysis and segmentation pipelines include CellProfiler and CellSegm~\cite{carpenter_cellprofiler:_2006, hodneland_cellsegm_2013}. The algorithm presented here represents a similar tool developed for our in-house use and, though not compared quantitatively to other pipelines, we hope it is similarly useful for a broader audience to better learn and implement such image analysis techniques. The code is freely accessible using the link above. We welcome any comments and feedback regarding the algorithm and its implementation.

\begin{acknowledgments}
L.R.B. was supported in part by NIH 1F31AR069977. L.R.B, L.J.B, and I.C. acknowledge support from NSF CCMI 1536463. Images were collected with user facilities in the Biotechnology Resource Center at Cornell University, with support from NIH S10RR025502, NYSTEM CO29155, and NIH S10OD018516. 

The authors would also like to thank members of the Cohen and Bonassar labs, especially E.D. Bonnevie, L.C. Chen, B. D. Leahy, M. L. Delco, and N. Diamantides.
\end{acknowledgments}

\bibliography{bibliography}

\end{document}